\documentclass[journal]{IEEEtran}
\usepackage[numbers]{natbib}

\ifCLASSINFOpdf
  \usepackage[pdftex]{graphicx}
  \graphicspath{{../pdf/}{../jpeg/}}
  \DeclareGraphicsExtensions{.pdf,.jpeg,.png}
\else
\fi
\usepackage{amsmath}
\usepackage{algpseudocode}

\usepackage{array}
\usepackage{fixltx2e}
\usepackage{url}

\usepackage{orcidlink}
\usepackage{amsfonts}
\newtheorem{definition}{Definition}
\usepackage{dsfont}
\usepackage{booktabs}

\begin{document}
%
\title{Exploration of the Rashomon Set Assists Trustworthy Explanations for Medical Data}
%
%

\author{Katarzyna~Kobylińska~\orcidlink{0000-0002-0292-4982},   
Mateusz~Krzyziński~\orcidlink{0000-0001-6143-488X},
Rafał~Machowicz~\orcidlink{0000-0002-2077-6610}, \\ 
Mariusz~Adamek~\orcidlink{0000-0002-1885-9257}, and
Przemysław~Biecek~\orcidlink{0000-0001-8423-1823}
\thanks{This work was financially supported by
HOMER NCN Sonata Bis-9 grant 2019/34/E/ST6/000522.}
\thanks{Katarzyna~Kobylińska is with the Faculty of Mathematics, Informatics, and Mechanics, University of Warsaw, 02-097 Warsaw, Poland (e-mail: kz277838@students.mimuw.edu.pl).}
\thanks{
Mateusz Krzyziński is with the Faculty of Mathematics and Information Science, Warsaw University of Technology, 00-662 Warsaw, Poland.}
\thanks{
Rafał Machowicz is with the Department of Hematology, Transplantation and Internal Diseases, Medical University of Warsaw, 02-091 Warsaw, Poland.}
\thanks{
Mariusz Adamek is with the Faculty of Medical Sciences in Zabrze, Medical University of Silesia, 41-800 Zabrze, Poland and the
Faculty of Health Sciences, Medical University of Gdańsk, 80-210 Gdańsk, Poland.}
\thanks{
Przemysław Biecek is with the Faculty of Mathematics, Informatics, and Mechanics, University of Warsaw, 02-097 Warsaw, Poland and the Faculty of Mathematics and Information Science, Warsaw University of Technology, 00-662 Warszawa, Poland.}
}

\maketitle

\begin{abstract}

The machine learning modeling process conventionally culminates in selecting a single model that maximizes a selected performance metric. However, this approach leads to abandoning a more profound analysis of slightly inferior models. Particularly in medical and healthcare studies, where the objective extends beyond predictions to valuable insight generation, relying solely on a single model can result in misleading or incomplete conclusions. This problem is particularly pertinent when dealing with a set of models known as \textit{Rashomon set}, with performance close to maximum one. Such a set can be numerous and may contain models describing the data in a different way, which calls for comprehensive analysis. This paper introduces a novel process to explore models in the Rashomon set, extending the conventional modeling approach. 
We propose the \texttt{Rashomon\_DETECT} algorithm to detect models with different behavior. It is based on recent developments in the eXplainable Artificial Intelligence (XAI) field. 
To quantify differences in variable effects among models, we introduce the Profile Disparity Index (PDI) based on measures from functional data analysis. 
To illustrate the effectiveness of our approach, we showcase its application in predicting survival among hemophagocytic lymphohistiocytosis (HLH) patients -- a foundational case study. Additionally, we benchmark our approach on other medical data sets, demonstrating its versatility and utility in various contexts. 
If differently behaving models are detected in the Rashomon set, their combined analysis leads to more trustworthy conclusions, which is of vital importance for high-stakes applications such as medical applications.
\end{abstract}


%
\IEEEpeerreviewmaketitle

\section{Introduction}
%
%
%
%
\IEEEPARstart{T}{he} machine learning (ML) process can serve at least two high-level purposes: predictive modelling and explanatory modeling focused on drawing conclusions about the modeled phenomena \cite{explainpredict}. Regardless of the purpose, process involves training a set of models from different families or a set of models for different hyperparameters and then selecting the best model from them. This methodology entails the selection of the model optimizing a specific metric, often presumed best to capture the data-generating process and its corresponding phenomenon. However, the highest-performing model may not align with domain knowledge. Moreover, it is plausible to have equally competent models, even with divergent data descriptions. Such multiplicity of good models was first conceptualized and analyzed by \citet{breiman}, who named such a phenomenon the \textit{Rashomon effect}. He described that one way to deal with this effect is to aggregate models from the Rashomon set, referring to the bagging \cite{bagging} procedure. Preparing the collection of multiple different models could improve predictive performance, as evidenced later also by ensembling methods \cite{ensembling}.

Considering the second purpose of the ML modeling process, related to inference about the problem studied, analyzing the whole collection of good models can be beneficial too. It provides a more comprehensive view of the investigated phenomenon, especially when the models from the Rashomon set describe the nature of data in a contrasting way. Although many studies have addressed the topic of Rashomon sets \cite{dong, rudin, Semenova}, the characterization of such a set of models is still an open problem \cite{challenges}. Selecting the most distinct models that explain the data differently is also an unresolved issue. 

In particular, the Rashomon effect poses a notable challenge in fields where insights from models are crucial, such as medical and healthcare applications. Despite this, many medical machine learning studies still prioritize predictive performance metrics over interpreting the model's prediction \cite{medical_challenges}. While some argue that divergences between models might be tolerable when clinically meaningful outcomes are robustly assessed \cite{accuracy_vs_explainability}, it is essential to recognize that numerous published studies on medical predictive models lack proper performance reporting and exhibit bias risks \cite{covid_models}. Thus, it is believed that model interpreting is essential  \cite{interpretability_pursuit} and methods aiming at model exploration \cite{predictiveModels}, often referred to as eXplainable Artificial Intelligence (XAI) techniques \cite{darpa-xai}, should be used in the modeling process.

\begin{figure*}[htb]
    \centering \includegraphics[width=0.8\linewidth]{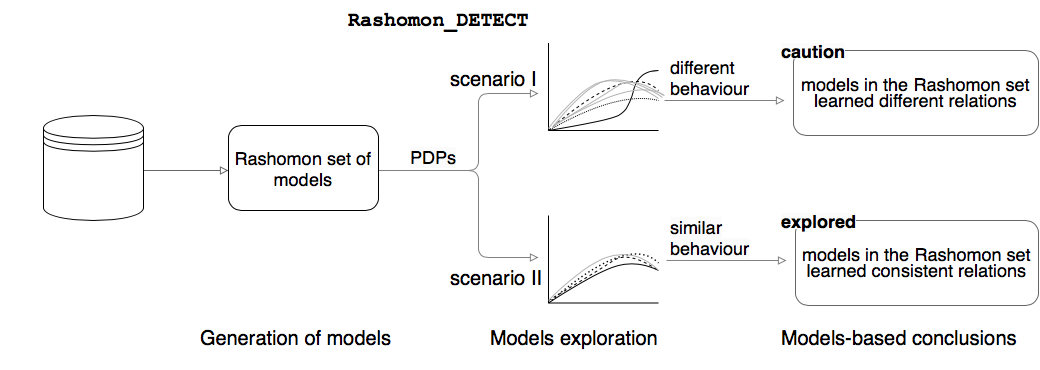}
    \caption{Schematic overview of reasoning with \texttt{Rashomon\_DETECT} algorithm. The ’caution’ field indicates potentially diverse models that ask for additional validation against domain knowledge. The ’explored’ field indicates the models with similar
relationships. This method can be used to provide a trustworthy model for a given
phenomenon. }
    \label{Workflow}
\end{figure*}

In this study, we aim to leverage XAI to address the Rashomon effect in machine learning by detecting the most different models from the considered set. For this purpose, we propose using explanation techniques that describe the dependence between the model's predictions and variable values (referred to as \emph{profiles}). Our approach centers around the step-wise \texttt{Rashomon\_DETECT} algorithm, designed to select a subset of $k$ models from a Rashomon set that exhibit near-identical predictive performance while manifesting extreme diversity in describing the studied phenomenon. The conceptual diagram is provided in Figure \ref{Workflow}.

A pivotal component of our algorithm is quantifying dissimilarities in variable effects between pairs of models. Although various existing measures are suitable for functional data \cite{fda_metrics}, we introduce a novel metric, the profile disparity index (PDI), designed to enhance this assessment. Our research demonstrates that, in several scenarios, the PDI provides a more comprehensive analysis of profiles compared to prevailing distance metrics.

Moreover, we demonstrate the application of the defined procedure to the medical data set on patients with hemophagocytic lymphohistiocytosis (HLH) -- a heterogeneous, potentially fatal syndrome of hyperinflammation. The data collection procedures were conducted in accordance with the guidelines and regulations of the Medical University of Warsaw Bioethics Committee, AKBE/37/15. This specific modeling task prompted the development of the proposed method. The case study showcases our technique's ability to capture variations in variable effects across models. This is particularly beneficial for selecting the model most compatible with domain knowledge, which is significant for personalized medicine and trustworthy modeling. Furthermore, we demonstrate the effectiveness and versatility of our method by benchmarking it on several other medical data sets, highlighting its adaptability and utility across diverse contexts. 

The aim of our study was to provide a more trustworthy approach to data analysis through a broader exploration of multiple models rather than relying solely on the performance measure values.
To facilitate the detection of different models, we have provided software that automates the process of determining the differences between models and selecting the best one. The software is available at \url{https://github.com/MI2DataLab/Rashomon-detect}.

\section{Related Work}
The concept of a collection of well-performing but different models known as the Rashomon set was initially introduced by Breiman \cite{breiman}. He demonstrated the problem using the example of five-variable linear regressions fitted on data with 30 variables. In this case, several models had similar residual sum-of-squares, but each used a different subset of variables. 
Later, this simple example led to the Rashomon set concept being analyzed from different perspectives by many researchers, also in the context of more flexible, complex models \cite{quartet}.

An approach based on the statistical learning theory is presented by
\citet{Semenova}. The authors explore the question of the existence of simple-yet-accurate models by presenting the Rashomon ratio gauge. It is the proportion of the Rashomon set's volume (number of models in the set) to the volume of the hypothesis space and can be used to check how complex the learning problem is. \citet{nevo} consider the minimal class of models similar in their prediction accuracy but not necessarily in their structure. The authors propose an algorithm to search the candidate models. \citet{theja} use Rashomon sets to prepare robust optimization problems to support decision-making on complex data. \citet{RashomonCapacity} introduce Rashomon capacity -- a metric to measure the predictive multiplicity in classification. \citet{EBEC} implement a methodology of leveraging the Rashomon effect of statistics in order to correct the explanations. Finally, \citet{Kissel} propose the Model Path Selection, an efficient method based on the forward selection to find accurate models from the Rashomon set. 

Several studies also explore the intersection of Rashomon sets with eXplainable Artificial Intelligence (XAI) methods. Primarily, globally applicable, model-agnostic techniques are favored, as they offer insights into a model's entire reasoning process without assuming anything about its structure. One such technique is Permutation Feature Importance (PFI), initially proposed for random forests \cite{randomForest} and then used by \citet{rudin} in a study on Model Class Reliance (MCR). The authors calculate the range of variable importance scores for each feature for models from the Rashomon set. MCR gives a more comprehensive description of importance than PFI, the method that takes only one model. 

The MCR concept has prompted further extensions. \citet{dong} do it by introducing the concept of the variable importance cloud and its visualization for models from the Rashomon set. \citet{RFMCR} extended computation of MCR to Random Forest models. \citet{xin} explore the Rashomon set of sparse decision trees, propose a technique for enumerating such set and investigate the variable importance using MCR. However, despite so many extensions, MCR was not used to delineate a subset of the most diverse models. It underscores the value of exploring alternative perspectives for characterizing models within the Rashomon set.

Specifically, answering the question of the variable importance leads to examining how the model uses the individual variables \cite{iema}. For this purpose, techniques that describe the dependence between the model's predictions and variable values (profiles) can be used. The most pertinent examples are Partial Dependence Plots (PDP) \cite{pdp} and Accumulated Local Effects (ALE) \cite{AlePlot}. Despite their popularity, to the best of our knowledge, these methods have not been studied yet in the context of Rashomon sets.

\section{Identifying disparate models from the Rashomon set through profile comparisons}

In this section, we describe the proposed novel approach of comparing models' reasoning and identifying the most different models of the Rashomon set. Our approach is specified through the step-wise \texttt{Rashomon\_DETECT} algorithm that makes use of profile dissimilarity measures, such as the newly proposed Profile Disparity Index PDI. We also underscore the connection between our approach and the potential to enhance established model development process frameworks. This emphasizes the profound impact our technique could have on modeling practices within the medical and healthcare domains.

\subsection{Notation and definitions}
In order to present the proposed solution, we introduce the following notation: $\mathbf{M} = [\mathbf{X} \; \mathbf{Y}] $ is a design matrix consisting of $n$ observations $\mathbf{X} = [\mathbf{x}_{1}, \mathbf{x}_2, \ldots, \mathbf{x}_n]^T$ of $p$ independent variables ($\mathbf{x}_i = [x^{(1)}_i, x^{(2)}_i, \ldots, x^{(p)}_i] \in \mathcal{X}$), and dependent variable vector $\mathbf{Y} \in \mathcal{Y}$. Let us consider a space of predictive models $\mathcal{F} = \{f \mid f: \mathcal{X} \rightarrow \mathcal{Y}\}$ -- these are measurable functions from $\mathcal{X}$ to $\mathcal{Y}$. Moreover, let $\mathcal{L}: (\mathcal{F} \times  \mathcal{X} \times \mathcal{Y}) \to \mathbb{R}$ denote the loss function used for the model performance evaluation. 

\begin{definition}[Reference model]
We call the reference model $f_{ref} \in \mathcal{F}$ one of the models selected from the space of predictive models according to a certain criterion in order to stand as a benchmark for the others. It can be a model that minimizes expected loss $\mathcal{L}$ over family of models $\mathcal{F}$, i.e., 

\begin{equation} 
f_{ref} = \mathrm{argmin}_{f \in \mathcal{F}} \mathbb{E}_{(\mathbf{X}, \mathbf{Y}) \sim  (\mathcal{X}, \mathcal{Y})} \mathcal{L}(f, \mathbf{X}, \mathbf{Y}),
\end{equation}
or any other reference model, such as the one currently used in practice.
\end{definition}

Having the reference model chosen, we can define the Rashomon set. 

\begin{definition}[Rashomon set]
For the given loss function $\mathcal{L}$, considered $\epsilon > 0$, and the reference model $f_{ref} \in \mathcal{F}$, the Rashomon set is 
\begin{align}
R_{\mathcal{F}}(\mathcal{L}, \epsilon, f_{ref})  = \{&f \in \mathcal{F} \mid \mathbb{E}\mathcal{L}(f, \mathbf{X}, \mathbf{Y}) \leq \\ & \mathbb{E}\mathcal{L}(f_{ref}, \mathbf{X}, \mathbf{Y})+\epsilon\} \subset \mathcal{F}, \nonumber
\end{align}
where expected values $\mathbb{E}$ are calculated with respect to the distribution of $(\mathcal{X}, \mathcal{Y})$.
\end{definition}

Let us observe that the loss function $\mathcal{L}$ can be any function that determines the predictive performance of the model (with possibly a corresponding change in the direction of inequality), e.g., mean squared error, the area under the ROC curve (AUC) or accuracy. In the experiments in this study, we consider a set of best models determined based on AUC values. We compare them using \emph{profiles}.

\begin{definition}[Profile]
For the given model $f \in \mathcal{F}$ and considered explanatory variable $j \in \{1, 2, \dots, p\}$, we call the \emph{profile} a function $g_{f}^{j}: \mathcal{D}_j \to \mathbb{R}$ that describes the relation between the values of the $j^{th}$ variable and model $f$'s predictions. 
\end{definition}

In practical considerations,
\begin{align}\mathcal{D}_j = [\mathrm{min}_{i \in \{1, 2, \dots, n \}} x_i^{(j)}, \mathrm{max}_{i \in \{1, 2, \dots, n \}} x_i^{(j)}].\end{align} There are several different ways of constructing and understanding profiles, of which the best known are Partial Dependence Plots (PDP) \cite{pdp}, Accumulated Local Effects (ALE) \cite{AlePlot}, or SHAP dependence plots \cite{lundberg2018consistent}. For PDP used in this study, the function $g_f^j(z)$ is defined as 

\begin{align} 
g_{f, PDP}^{j}(z) = \mathbb{E}_{\mathcal{X}^{(-j)}}[&f(X^{(1)}, \dots, X^{(j-1)}, \\ &z, X^{(j+1)}, \dots, X^{(p)})], \nonumber
\end{align}
which is the expected value of model predictions for a fixed value of the $j^{th}$ variable at $z$ over the joint distribution of $\mathcal{X}^{(-j)}$ -- all explanatory variables other than the $j^{th}$ one. It can be easily estimated as:
\begin{align}
\widehat{g_{f, PDP}^{j}} (z) = \frac{1}{n} \sum_{i=1}^{n} f(x^{(1)}_i, \dots, x^{(j-1)}_i,  z, x^{(j+1)}_i,\dots, x^{(p)}_i).
\end{align}
Let us emphasize that this is a model-agnostic method and can therefore be used for all predictive models that could differ greatly in their mathematical formalism or structure. Moreover, as each of the methods has its strengths and weaknesses, the choice of the particular technique used in the process belongs to the researcher; e.g., ALE takes into consideration the correlation between the variables.

\subsection{Measuring profile differences}
When assessing model reasoning based on profiles, various approaches can be utilized, employing functions that compare two profiles and quantify the level of dissimilarity between them using methods like distance metrics.

In functional data analysis \cite{fda_metrics}, one common option is the $L^2$ metric (Euclidean distance) between functions. It measure distance between profiles $g_{f_1}^j, g_{f_2}^j: \mathcal{D}_j \to \mathbb{R}$ calculated for models $f_1$, $f_2$ and explanatory variable $X^{(j)}$ as
\begin{align}
d_{L^2}(g_{f_1}^j, g_{f_2}^j) = \sqrt{\int_{\mathcal{D}_j} \left( g_{f_1}^j(z) - g_{f_2}^j(z) \right)^2 \mathrm{d}z}.
\end{align} 

However, in many cases, the similarity of profiles depends on their shapes rather than their predictive values. To capture this difference with emphasis on shape, a measure that accounts for monotonicity can be employed. The $L^2$measure between first-order derivatives (a semi-metric) serves this purpose:
\begin{align}
d_{L^2,der}(g_{f_1}^j, g_{f_2}^j) = \sqrt{\int_{\mathcal{D}_j} \left( \frac{\partial}{\partial X^{(j)}} g_{f_1}^j(z) - \frac{\partial}{\partial X^{(j)}} g_{f_2}^j(z) \right)^2 \mathrm{d}z}.
\end{align} 

In the domain of linear models often used in the medical context, a scenario arises where profiles for the same variable may show slight differences due to different regularization strengths, often maintaining an unchanged monotonicity relationship (the sign of the associated coefficient remains consistent). Taking this notion further, we introduce a new measure that, by disregarding the rate of changes in profiles, can effectively handle minor fluctuations in coefficients within linear models and even cope with oscillatory profiles observed in more elastic models. This measure focuses on discerning whether one profile ascends while the other descends. This divergence in directional behavior signifies distinct predictive dependencies within the models under examination. We refer to this metric as the Profile Disparity Index (PDI). Its formulation is as follows:

\begin{align}
PDI(&g_{f_1}^j, g_{f_2}^j) = \frac{1}{\sup \mathcal{D}_j - \inf \mathcal{D}_j} \\ & \int_{\mathcal{D}_j}\left[\mathrm{sgn} \left( \frac{\partial}{\partial X^{(j)}} g_{f_1}^j(z) \right) \neq \mathrm{sgn} \left( \frac{\partial}{\partial X^{(j)}} g_{f_2}^j(z) \right) \right] \,\mathrm{d}z. \nonumber
\end{align}

Naturally, it can be estimated based on empirical profiles $\widehat{g_{f_1}^j}, \widehat{g_{f_2}^j}$ as 
\begin{align}
    \widehat{PDI}
    (
    \widehat{g_{f_1}^j}, 
    \widehat{g_{f_2}^j}
    ) = 
    \frac{1}{m}\sum_{i=1}^{m} \mathds{1}
    [ 
    & \mathrm{sgn} (\mathrm{der}(\widehat{g_{f_1}^j})[i]) \neq \\
    & \mathrm{sgn} (\mathrm{der}(\widehat{g_{f_2}^j})[i]) ], \nonumber 
\end{align}
where $\mathrm{der}(\widehat{g_{f_k}^j})$ is the vector of derivatives determined at $m$ consecutive points of the profile for the $k^{th}$ model and $\mathrm{der}(\widehat{g_{f_k}^j})[i]$ is the $i^{th}$ element of this vector. Note that profile values are, in practice, determined for selected values of an explanatory variable, e.g., for arguments from a uniform grid or all values occurring in the learning sample. Whereas derivatives can be determined using Generalized Orthogonal Local Derivative (GOLD) method \cite{gold}. 

The interpretation of the empirical version of PDI is very intuitive. Namely, it is the percentage of disparity between two profiles and takes values in the range $[0,1]$. The value zero indicates that two curves are the same, whereas one indicates that two curves differ radically. 

In scenarios involving categorical variables, the use of derivatives is not feasible. To address this limitation, we employ vector distances as an alternative method for measuring dissimilarities in such cases. This approach helps bridge the gap in our methodology and ensures a comprehensive analysis across diverse variable types.

\subsection{\texttt{Rashomon\_DETECT} algorithm}
Identifying the spectrum of models with extreme behaviors from a Rashomon set may be of critical importance and is a major step towards exploring a group of models rather than analyzing a single one. Hence, we propose the \texttt{Rashomon\_DETECT} algorithm that addresses this issue as its objective is to select the $k$ most different models. It operates on the set of the created models $\hat{\mathcal{F}}$ and is based on the calculation of the average dissimilarities between profiles for pairs of models. These dissimilarity values are computed per variable and then averaged. In cases where models possess distinct variable sets, any missing profiles are filled using a constant function $\mathbf{0}$. The algorithm starts by adding to the resulting set the reference model, which is most often the one that minimizes the given loss function. In subsequent steps, it augments the result set with the most dissimilar model based on a chosen measure (average value of dissimilarity for already selected models is used). The simplified algorithm is as follows:

\begin{algorithmic}[1]
\Statex
\Require $\hat{\mathcal{F}}$ -- family of the created models, $\mathcal{L}$ -- loss function, $\epsilon > 0$, $k\geq2$ -- number of models to detect, $g$ -- method for estimating profiles, $d$ -- measure for quantify dissimilarities between profiles
\Ensure $\mathcal{R}$ s.t. $|\mathcal{R}| = k$ -- subset of the most different models from Rashomon set
\State $f_{ref} \gets \mathrm{argmin}_{f \in \mathcal{F}} \mathcal{L}(f)$ \Comment{start with the provided reference model or find it}
\State $R \gets \{f \in \hat{\mathcal{F}} \mid \mathcal{L}(f) \leq \mathcal{L}(f_{ref})+\epsilon\}$ \Comment{determine the models to be considered}
\If{$k \geq |R|$} \Comment{not enough models for given $k$}
    \State $\mathcal{R} \gets R$
\Else
    \State $\mathcal{R} \gets \{f_{ref}\}$
    \State $R \gets R \setminus \{f_{ref}\}$
    \While{$|\mathcal{R}| \leq k$}
        \State $f^* \gets \mathrm{argmax}_{f \in R} \;
        \frac{1}{|\mathcal{R}| \cdot p} \sum_{i=1}^{|\mathcal{R}|} \sum_{j=1}^{p} d( \widehat{g_{f_{\mathcal{R}_i}}^j}, \widehat{g_{f}^j})$ \Comment{select model with the biggest average dissimilarity to models in $\mathcal{R}$}
        \State $\mathcal{R} \gets \mathcal{R} \cup \{f^*\}$
        \State $R \gets R \setminus \{f^*\}$
    \EndWhile
\EndIf
\State \Return $\mathcal{R}$
\Statex
\end{algorithmic}



In addition to the designated set of the most different models, the algorithm can yield calculated profiles and measures of the dissimilarity between them, i.e., between individual profiles for single variables and the averaged values used to build the resulting set. These results are to be used for further analysis of models, e.g., obtained values can aid in clustering models into distinct groups.

However, in the case of a large set of models, it may be valuable to use a simplified form of the algorithm, which works greedily by including subsequent models only based on comparison with the most recently added one.

\subsection{Relevance to modeling process}

Many machine learning procedures have been proposed in order to systematize the modeling process. Their goal is to find the best model describing the studied phenomenon. The most well-known framework, called Cross-industry Standard Process for Data Mining (CRISP-DM), was introduced by \citep{Chapman}. Many authors have published an analysis of that process, e.g., \citet{Crispdm} prepared a comprehensive analysis with tests of that methodology. 
In \cite{predictiveModels}, the authors propose the life-cycle of a predictive model as an iterative 5-step process: (1) data preparation, (2) data understanding, (3) model selection, (4) model audit, (5) model delivery. 

Our approach enhances these frameworks, particularly in the context of medical applications and healthcare decision-making. We propose to broaden the process via consideration of a set of good models instead of only one model. By validating and comparing the $k$ most distinct models from the generated set, we can arrive at a more nuanced and reliable conclusion. 

When the most diverse models do not significantly differ in their explanations of the data, this outcome can foster greater trust in the analytical results. Confirming consistent model behavior across the set assures the reliability of the entire analytical process. On the contrary, if evident disparities are apparent, the model selected for final delivery need not be the one that optimizes the loss function. Alternative selection criteria can emerge depending on the clinical scenario and the characteristics observed. These might include models that align more closely with domain knowledge, excel in handling atypical patients (outliers), or exhibit enhanced stability and robustness. Furthermore,  we recognize the potential to provide a collection of models, each presenting diverse perspectives on the analyzed phenomenon. Such an ensemble of models can assist clinicians in selecting the model most appropriate for a given patient, tailoring the medical approach to individual needs.

\section{Experiments}\label{sec3}

\begin{figure*}[htb]
    \centering \includegraphics[width=0.9\linewidth]{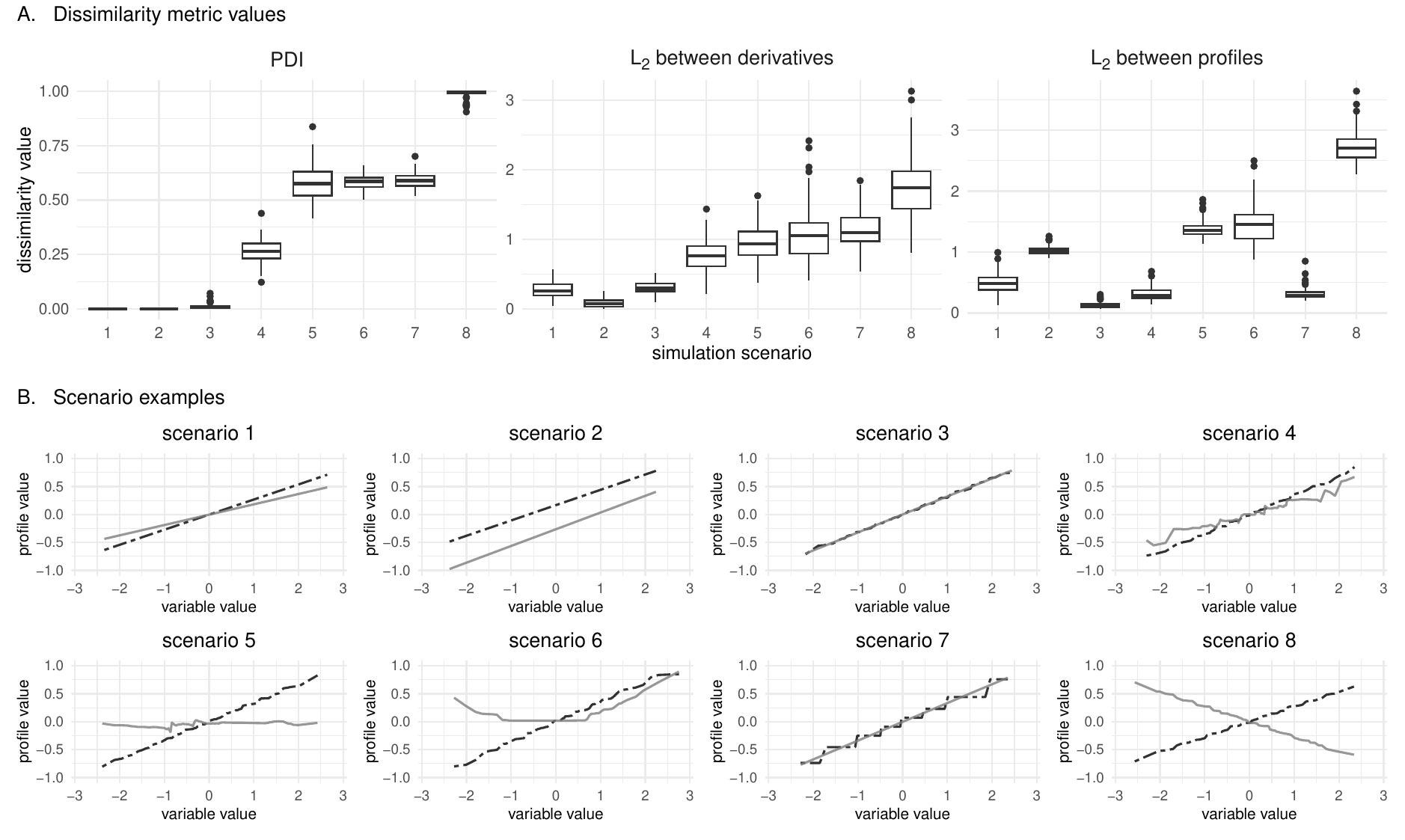}
    \caption{(Panel A) Distributions of dissimilarity metric values for the analyzed scenarios. Higher values reflect increased profile dissimilarity based on the corresponding measure. Note that the y-axes vary due to each measure's distinct theoretical value ranges. (Panel B) Example pairs of profiles generated for each examined scenario.}
    \label{metrics}
\end{figure*}

This section delineates the experimental procedures conducted to assess the efficacy of the proposed solution. The first experiment serves to compare different measures of profile dissimilarities in various synthetically generated scenarios. Subsequently, the second experiment employs the real-world HLH data set to execute the \texttt{Rashomon\_DETECT} algorithm and to conduct an in-depth analysis of the resultant outcomes. This showcases a complete use case of the proposed approach. Lastly, the third experiment comprehensively evaluates the \texttt{Rashomon\_DETECT} algorithm's performance and adaptability when applied to other medical data sets.

\subsection{Evaluating profile dissimilarity measures on synthetic data}
\label{measures_exp}
In this experiment, we aim to demonstrate the behavior of the discussed measures, including the newly introduced PDI, when applied to eight distinct scenarios involving pairs of profiles with varying relationships. Notably, while PDI is bounded within the interval [0, 1], the other measures, though non-negative, lack an upper limit and can assume arbitrary values from the unbounded interval $[0, \infty)$, depending on the values of the profiles under consideration. Thus, a direct comparison of measure values across different scenarios cannot be conducted. Instead, we analyze the values across diverse scenarios and their relative relationships (rankings) within each measure.

Moreover, to make the values of both $L^2$ measures comparable between the different scenarios, we generate all profiles within the range of [-1, 1]. Furthermore, the domain for each profile is equivalent, as the synthetic variable in every scenario is sampled from a standard normal distribution $N(0, 1)$. In each scenario, we generate 100 pairs of profiles that vary due to random noise. This generation process employs regression machine learning models tailored for this purpose, ensuring the plausibility of resulting profiles that include oscillations and noise. 

The experiment showcases a variety of scenarios illustrating different positional relationships between profiles. For instance, scenario 1 involves profiles generated from linear regression models, where one utilizes only the variable under analysis while the other incorporates a second variable. Scenario 4 mirrors this with more flexible random forest models. On the other hand, scenario 7 juxtaposes a simple decision tree model with a linear model applied to the same data. However, note that the primary focus is to elucidate varying profile relationships rather than exhaustive model scenarios.

The experiment results are visualized in Figure \ref{metrics}. Panel A displays the distribution of metric values across scenarios, while Panel B provides examples of the analyzed scenarios. Generally, the PDI measure effectively discriminates between scenarios where profiles exhibit similar relationships (e.g., scenarios 1-4) and those where substantial differences are apparent (e.g., scenarios 5-6 and 8). The remaining measures show less distinct values between these two groups. Notably, median values of each measure peak in scenario 8, portraying distinctly inverse relationships. For PDI, most values reach the maximum of 1, with minor deviations attributed to noise present in profiles.

The proposed PDI measure consistently attains a value of~0 for scenarios 1 and 2, representing smooth profiles with a consistent monotonic relationship (both profiles increase). This holds regardless of whether profiles intersect or vary in inclination. In contrast, the semi-metric between derivatives yields notably different values for these scenarios due to the distinct rates of profile value increase with variable changes. The PDI values are also close to 0 for nearly identical profiles with slight noise (scenario 3), where the conventional Euclidean measure also performs well. 

In scenario 7, a stepped profile is compared to a linear one. Although akin to scenario 3 in exhibiting similar trends, the non-linear profile features larger oscillations around the line. This leads to elevated PDI and second derivative-based metric values compared to most other scenarios, primarily due to intervals of constant values yielding derivatives of 0. In contrast, the Euclidean metric perceives this situation differently, yielding relatively low values. A comparable but less pronounced relationship is observed in scenario 4. Consequently, the Euclidean measure may be more appropriate for profiles with large oscillations (e.g., high noise, stepped functions resulting from single shallow trees).

The experiment underscores the absence of one \emph{best} measure for assessing profile differences. Instead, it highlights that each measure has strengths and limitations across different scenarios. While the PDI measure excels in most scenarios, it is crucial to recognize that no measure guarantees optimal performance in every context, especially in intricate medical decision-making scenarios.

\subsection{Real-world case study -- hemophagocytic lymphohistiocytosis (HLH)}

We present the use of the \texttt{Rashomon\_DETECT} algorithm and the analysis of its results for real-world medical data. This case study was the origin and motivation for developing the method described. It concerns the classification task on the medical data set comprising 19 variables characterizing 101 adult patients with hemophagocytic lymphohistiocytosis (HLH; hemophagocytic syndrome). HLH is a rare syndrome of extreme hyperinflammation, which, if untreated, is almost uniformly fatal. Even with optimal treatment mortality rate is high. However, if initial hyperinflammation is silenced and then controlled, the prognosis gets much better with time – with six-month survival being a good predictor of treatment success. HLH may be triggered by various factors (malignancies, infections, autoimmune disorders). Moreover, the degree to which different organs and parameters are affected varies between patients. This heterogeneity makes it very difficult to predict the outcome. The main goal of the prediction task is to predict whether the patient survives six months after diagnosis. 



A benchmark of machine learning algorithms aimed at predicting the survival of patients over a six-month period has been conducted. We assembled a diverse set of models, including support vector machines (SVM), decision trees, gradient boosting machines (GBM), and random forests (RF). The data set was partitioned into distinct training and testing subsets, facilitating the evaluation of model performance using the AUC metric. The hyperparameter tuning was performed across all methods to identify models yielding the highest AUC values. The most outstanding performance emerged from one of the Random Forest models, achieving an AUC of 0.81 on the test data set, thus establishing itself as our reference model. Notably, five additional Random Forest models and four Gradient Boosting models, with AUC values exceeding 0.77, were also integrated into the analysis. Conversely, each SVM model and every decision tree model demonstrated comparatively lower performance metrics, rendering them excluded from the Rashomon set. 

\begin{figure}[!t]
\centering
\includegraphics[width=3.4in]{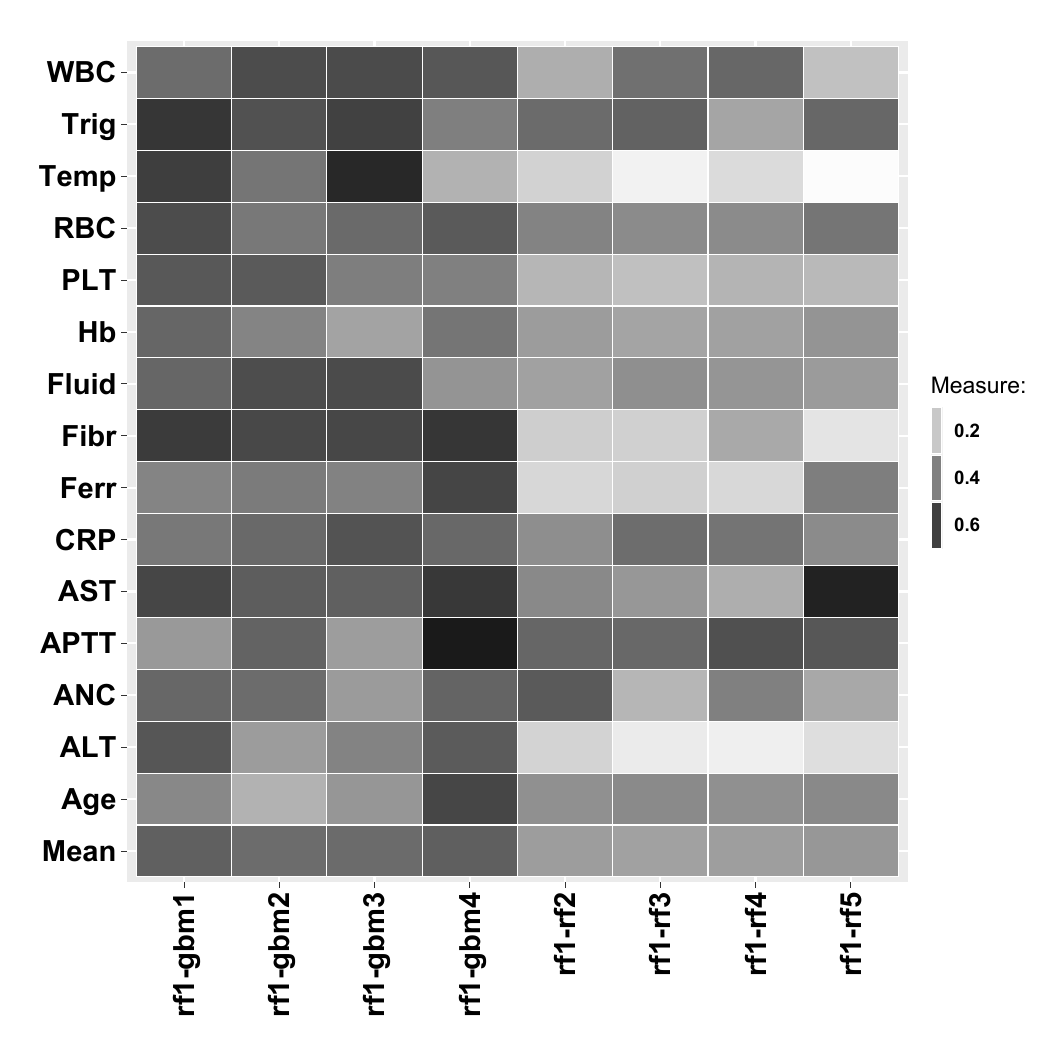}
\caption{The comparison of mean PDI measure for pairs: \texttt{rf1} and each analyzed model. The darker the field is, the lower the measure is, meaning bigger differences between models. The following abbreviations refer to feature names: \texttt{ALT} (Alanine aminotransferase), \texttt{ANC} (Absolute Neutrophil Count), \texttt{APTT} (Activated Partial Thrombin Time), \texttt{AST} (Aspartate aminotransferase), \texttt{Age}, Bilirubin, \texttt{CRP} (C-Reactive Protein), \texttt{Ferritin}, \texttt{Fibronectin}, \texttt{Fluid}, \texttt{Hb} (Hemoglobin), \texttt{PLT} (Platelet Count), \texttt{RBC} (Red Blood Cell Count).}
\label{heatmap}
\end{figure}

We want to detect the three most distinct models from the Rashomon set. The determination of the suitable 
k value depends on the size of this set. 
The determination of an appropriate value for
$k$ depends on the size of the Rashomon set. Given that the Rashomon set comprises only nine models, we opted for the square root of this quantity. Our objective was to identify three models that exhibit the most disparate outcomes.
The random forest model, called \texttt{rf1}, is the reference model with the highest AUC.
Subsequently, for each pair, the winner model, and each model from the Rashomon set, we have computed the PDI measures. Interestingly, all the combinations involving \texttt{rf1} with GBM models displayed greater measures than those involving other Random Forest models, as visually evident in the heatmap shown in Figure \ref{heatmap}. The model demonstrating the most substantial mean measure value across the combinations is Gradient Boosting Model, called \texttt{gbm1}. Consequently, we compiled the measures for \texttt{gbm1} alongside the remaining models. In the second step of our analysis, the prevailing model is identified as \texttt{gbm4}. 
 
\begin{figure}[!t]
\centering
\includegraphics[width=3.6in]{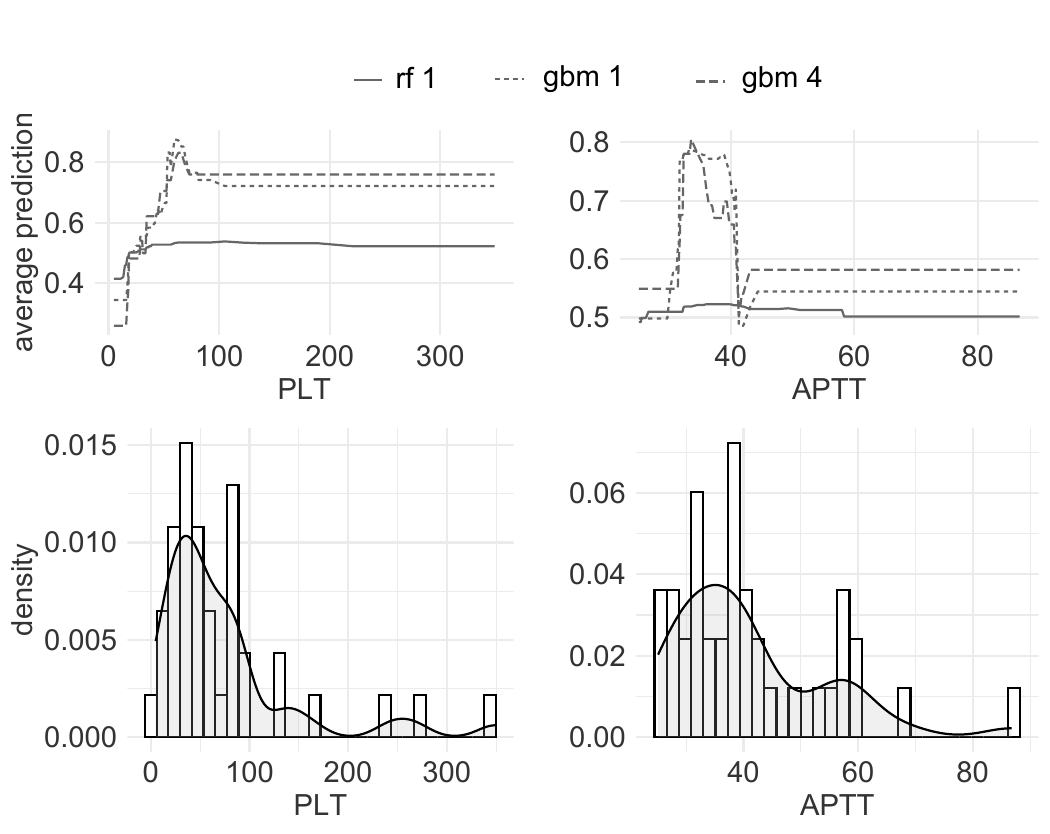}
\caption{Partial Dependence Plots for the detected variables with the highest PDI values.}
\label{PDPhist}
\end{figure}

A thorough examination of the profiles for the selected models was conducted. The highest PDI measure values (exceeding 50\%) are observed for variables: \texttt{APTT} (Activated Partial Thrombin Time) and \texttt{PLT} (Platelet Count) - parameters associated with different blood coagulation mechanisms. To provide visual insight into these variables, Figure \ref{PDPhist} presents the Partial Dependence Plots with the corresponding histograms. The PDPs confirm the disparities among the profiles that were originally identified by means of the PDI measure. 

Analyzed models differ in approach to \texttt{PLT} and \texttt{APTT}. Low \texttt{PLT} may result in bleeding, which can even cause death. This risk is mostly present in values below 20 $G/l$, which is reflected in all three models. In higher values, \texttt{rf1} model quickly reaches a plateau, while both \texttt{gbm1} and \texttt{gbm4} models continue to discern differences even among these higher values. Those differences are more subtle and reflect rather overall HLH activity than bleeding risk – such effect may not directly affect the prognosis and despite neglecting it \texttt{rf1} still obtained a higher AUC. Moreover, \texttt{gbm1} and \texttt{gbm4} models show their superiority in analyzing \texttt{APTT}. They clearly show that patients with normal \texttt{APTT} coagulation time have a higher probability of survival ($70 - 80\%$), and this rapidly decreases to less than $60\%$ when \texttt{APTT} is prolonged (above $40$s) – what may reflect the higher risk of life-threatening bleeding. 

The presented application of the \texttt{Rashomon\_DETECT} algorithm to medical data shows that several almost accurate models can describe the nature of the data in different ways. The procedure enables safer inference about medical problems because it takes into consideration a few models with the most different perspectives on the phenomenon. It allows the researcher or practitioner to choose one best-behaving model according to domain knowledge or use several models with different but reasonable reasonings.

\subsection{Benchmark on real-world medical data sets}

The benchmark study aims to evaluate the effectiveness of \newline the \texttt{Rashomon\_DETECT} algorithm applied to machine learning models computed based on diverse medical data sets. The considered collection of data sets covering different medical problems comes from the study on SeFNet. Detailed information on the data sets can be found in \cite{2023sefnet}.

For each data set, a similar process of searching for optimal models is evaluated. The benchmarking approach comprises the following steps: 5-time cross-validation during the hyperparameter tuning and creating the Rashomon sets according to the mean result from cross-validation. The step of hyperparameter tuning involves exploring a parameter grid, focusing on two classes of machine learning models: random forest and gradient boosting models. The Rashomon sets encompasses all models whose mean Area Under the Curve (AUC) from the 5-time cross-validation is not lower than the predetermined tolerance. 

The focal point of our investigation lies in the \texttt{Rashomon\_DETECT} algorithm aiming to determine its efficacy in identifying the most diverse profiles within the considered sets of models. Table \ref{table_benchmark} presents the essential details concerning the benchmark outcomes. 
The procedure of \texttt{Rashomon\_DETECT} has been executed thrice on each data set, each time with a distinct measure.

\begin{table}[!t]
\caption{Average measure values for pairs formed from the three most distinct models obtained in the benchmark.}
\label{table_benchmark}
\centering \begin{tabular}{cccccc}
  \toprule 
 {\bfseries Measure} &  ~{\bfseries COVID}~&  ~{\bfseries PIMA}~ & ~{\bfseries ILPD }~ & ~{\bfseries Heart }~   \\  \midrule

 PDI &  0.42 & 0.37& 0.42 & 0.19  \\
 &  0.39 & 0.35 & 0.32 & 0.16  \\ 
 &  0.31 & 0.13 & 0.16 & 0.16  \\
 \midrule 
 $L^2$ between & 0.08 & 0.1 & 0.7 & 0.08  \\
  profiles  & 0.06 & 0.09 & 0.57 & 0.06  \\
      & 0.03 & 0.04 & 0.07 & 0.06  \\
     \midrule
  $L^2$ between & 0.13 &0.09& 0.06 &0.08 \\
    derivatives  & 0.1 & 0.07& 0.05&0.06 \\
& 0.07 & 0.04 & 0.02 &0.06 \\
\bottomrule
\end{tabular}
\end{table}

\begin{figure*}[htb]
    \centering \includegraphics[width=1\linewidth]{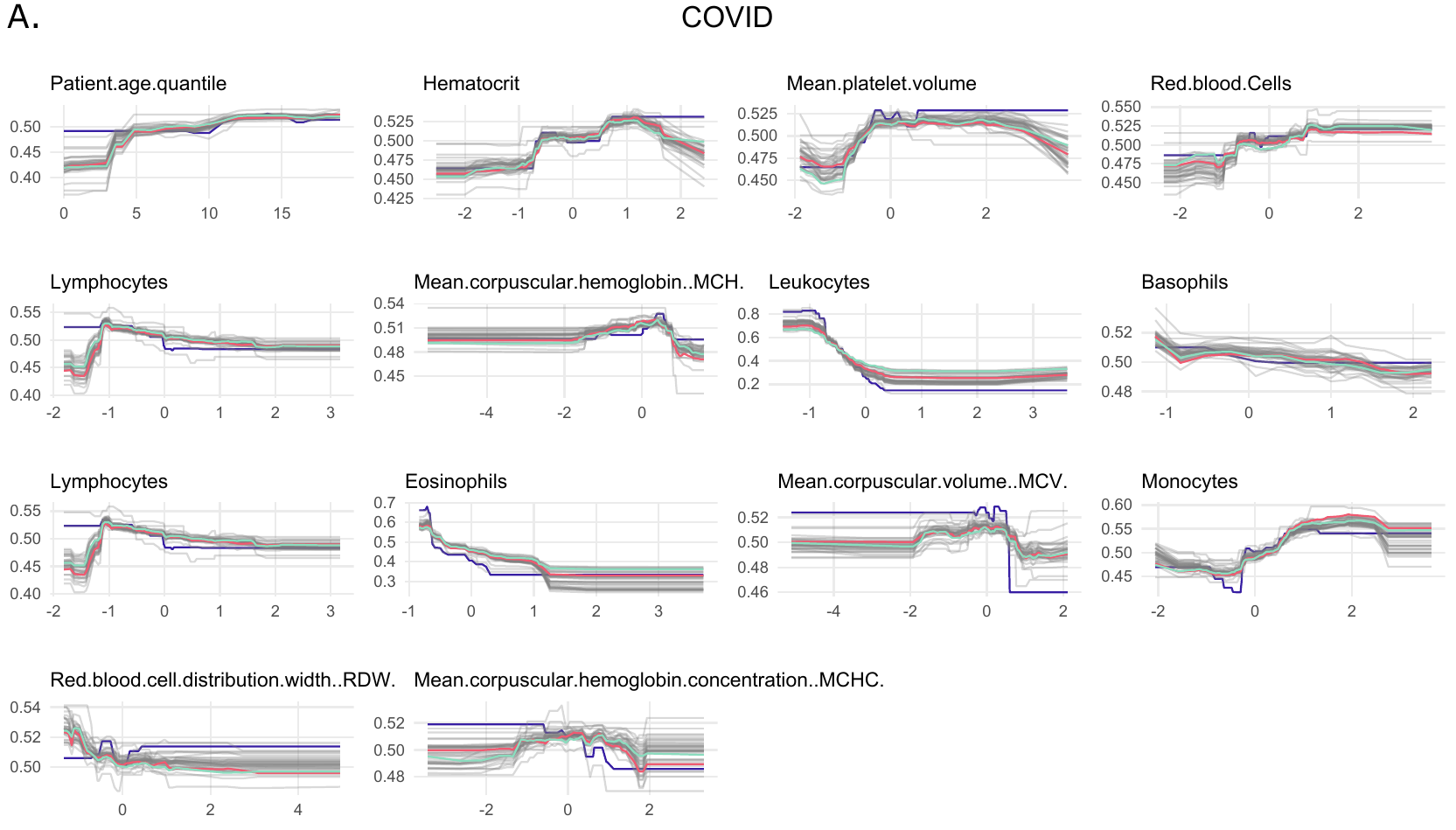}
    \includegraphics[width=1\linewidth]{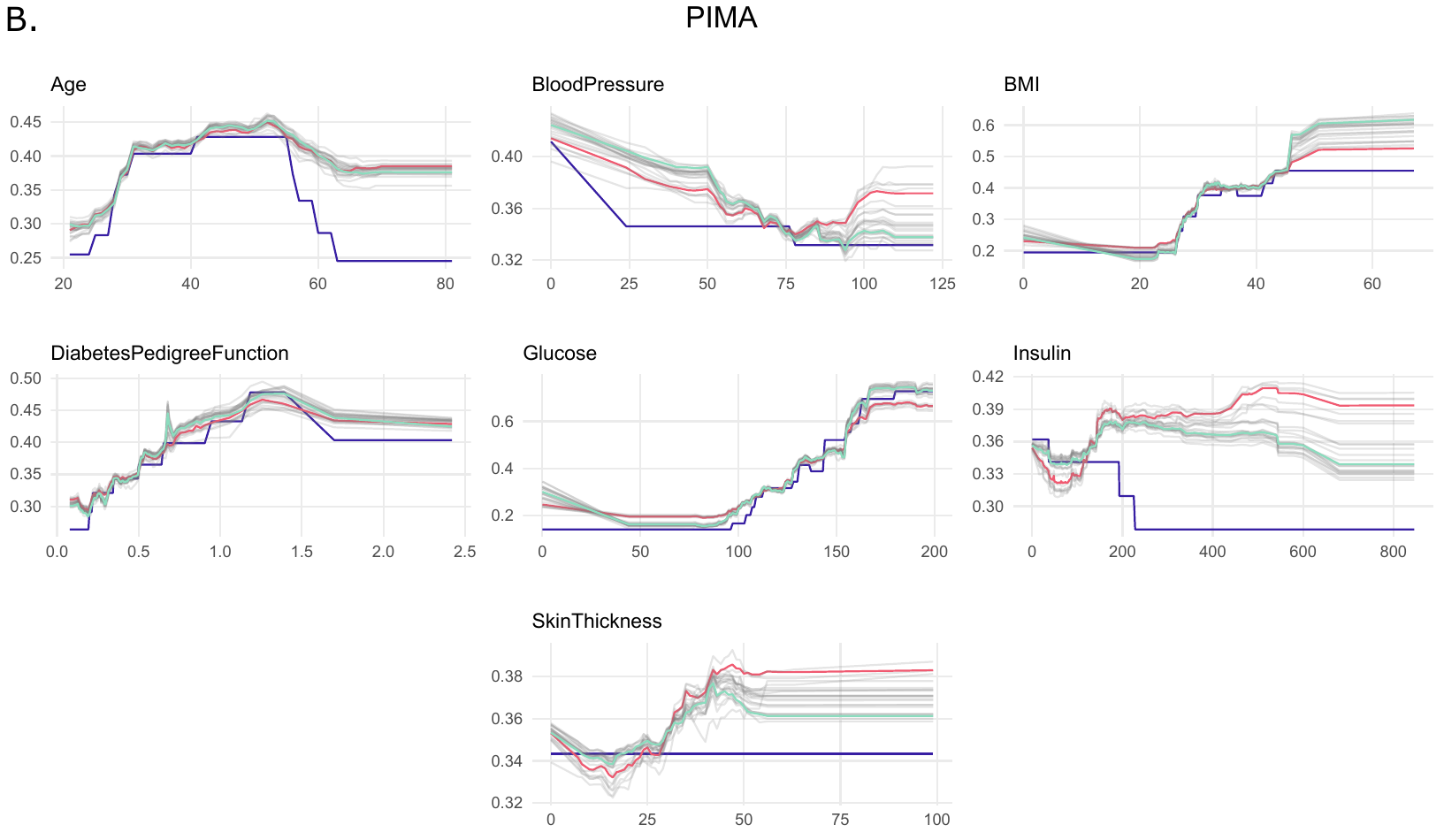}
    \caption{Partial Dependence Plots for continuous variables in medical data sets: (A) COVID data set, (B) PIMA data set. The lines represent PDPDs for various models from the Rashomon set. The three most distinct models detected for each data set are highlighted in red, green, and blue. The choice of metric in the \texttt{Rashomon\_DETECT} process was based on profile shapes and scenarios analyses outlined in Section~\ref{measures_exp}.}
    \label{PDP1}
\end{figure*}

\begin{figure*}[htb]
    \centering
    \includegraphics[width=1\linewidth]{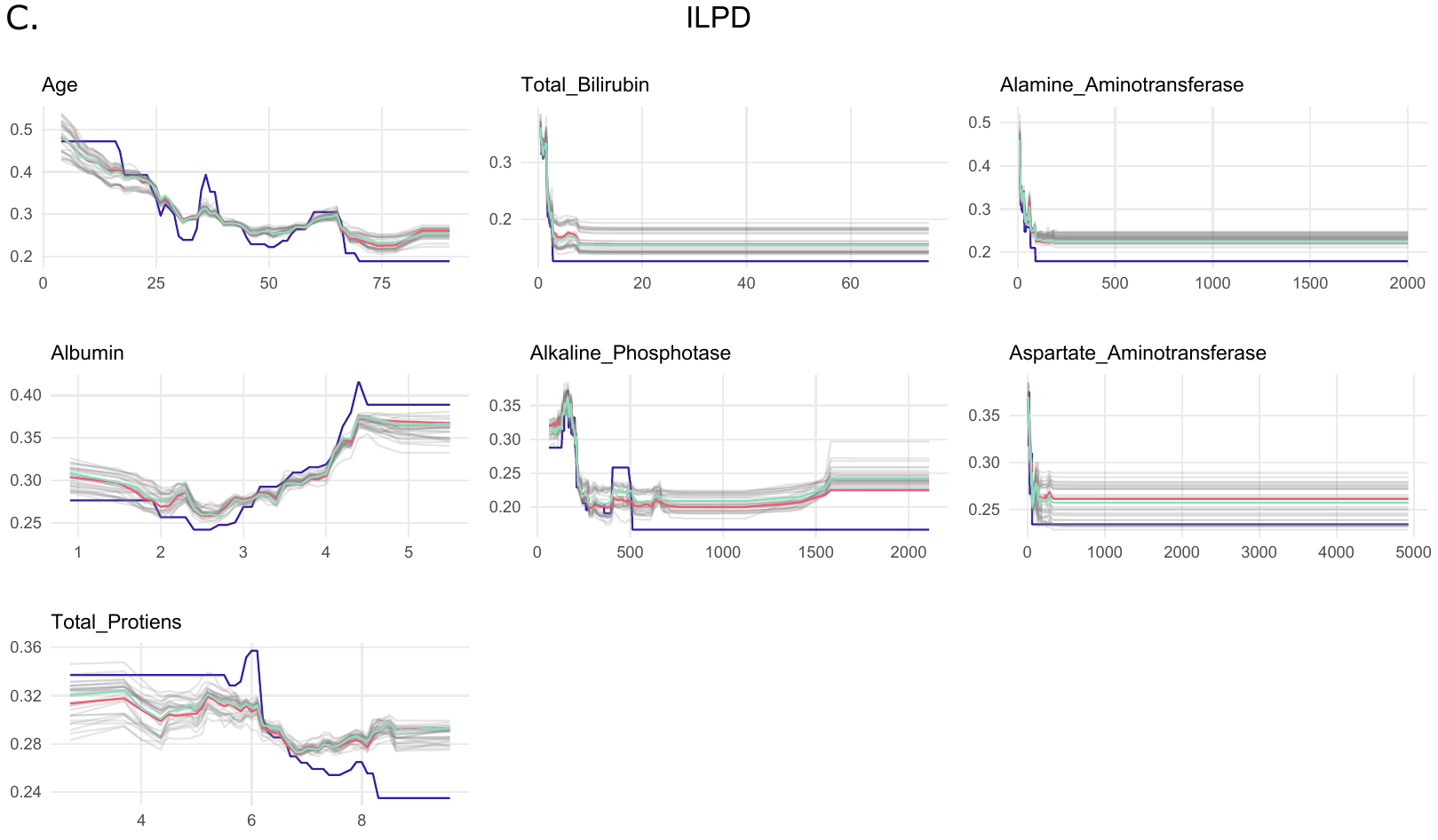}
    \includegraphics[width=1\linewidth]{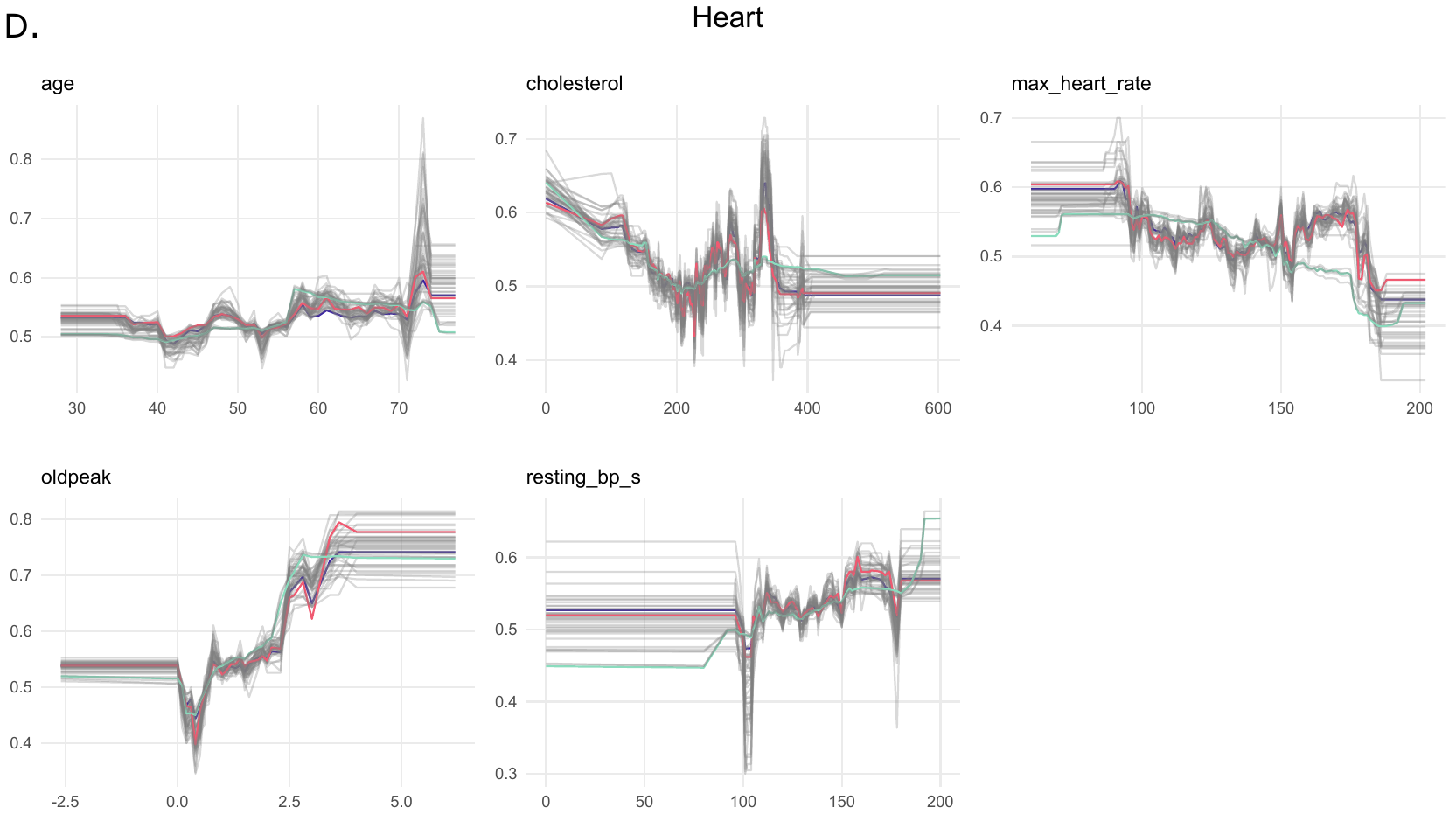}
    \caption{Partial Dependence Plots for continuous variables in medical data sets: (C) ILPD data set, (D) Heart data set. The lines represent PDPDs for various models from the Rashomon set. The three most distinct models detected for each data set are highlighted in red, green, and blue. The choice of metric in the \texttt{Rashomon\_DETECT} process was based on profile shapes and scenarios analyses outlined in Section~\ref{measures_exp}.}
    \label{PDP2}
\end{figure*}

The algorithm's effectiveness is measured by identifying actual disparities in profiles. We provide a confirmation by plotting individual profiles for all models and emphasizing by color the models in which the algorithm has indicated significant differences. 
Figures \ref{PDP1} and \ref{PDP2} present the Partial Dependence Plots associated with each model, serving as validation for the accuracy of the outcomes achieved by the \texttt{Rashomon\_DETECT} algorithm. 
The findings, as presented in this manner, confirm that the algorithm predominantly identifies the models with the most dissimilar profiles as the most distinct ones.
As detailed in Section \ref{measures_exp}, the findings underscore that diverse measures exhibit varying degrees of compatibility with individual data sets. Therefore, the plotted profiles presented in Figures \ref{PDP1} and  \ref{PDP2} illustrate the algorithm's result using the most compatible measure for each data set. 

For the Heart data set, wherein the majority of profiles exhibit oscillations aligning with scenario 4 and scenario 7, the Euclidean measure is adopted. The profiles derived from the PIMA data set largely resemble scenario 6, leading to the application of the PDI measure. Furthermore, the profiles associated with the ILPD and COVID data sets exhibit similarities to scenarios 6 and 4, leading to the application of the PDI measure.

In conclusion, our research underscores the importance of considering multiple explanations within Rashomon sets. By employing the disparity measure and \texttt{Rashomon\_DETECT} algorithm, we can reveal the most diverse and distinct models, enriching our comprehension of the underlying complexities in predictive modeling. This approach serves as a critical step towards advancing the reliability and interpretability of machine learning models in various domains. As an illustration, the models identified for the PIMA data set exhibit notable discrepancies in their treatment of age and insulin.

In the context of this benchmark, it is essential to acknowledge that various factors influence the composition of the final Rashomon set. At the outset of the modeling process, certain assumptions are made, such as the types of models to be tested, the parameter grids explored, or the chosen AUC tolerance. The primary objective of this benchmark is not solely to identify the largest possible Rashomon set.  Instead, we aim to demonstrate the existence of Rashomon sets comprising models with similar performance but distinct underlying explanations. In essence, we seek to showcase the diversity of plausible models that can achieve comparable results and the method how to find such models.

\section{Discussion}

This study introduces a novel process for exploring models from the Rashomon set. This approach can be applied to any machine learning problem to provide a broader perspective on explaining the data. The selection and analysis of machine learning models from the Rashomon set can give more trustworthy results than analyzing a single model alone. This is particularly crucial in high-stakes decision domains, such as medical problem modeling.
Especially when there are models that perform almost as well, their behaviors should be compared. Rather than comparing the models directly, we propose to compare the Partial Dependence Profiles. If the interpretations of the models differ, the results and final decision should be reinforced, for example, through domain knowledge or other XAI techniques. Moreover, such a broader view gives more trust in the final model. 
In addition, we show the possibility of aggregating PDP by calculating the PDI measure, which facilitates the interpretation of the results. 

However, it is important to acknowledge the algorithm's limitations. For instance, its outcomes are significantly influenced by the chosen metric in \texttt{Rashomon\_DETECT} or the chosen $\epsilon$ when constructing the Rashomon set. 

Despite the described findings, some opportunities for further exploration in this field persist. First, the process could be effectively applied using alternative eXplainable Artificial Intelligence methods. Moreover, calculating the similarity for the pair of models could be reinforced by, for example, the weighted average of the PDI or the implementation of a sliding window.
Additionally, investigating scenarios with a high number of variables within the data set could bring valuable insights.
An additional aspect worth considering is the comparative analysis between the outcomes of our algorithm and the results received from logistic regression models. Such a comparative study would provide deeper validation and understanding of our approach.

\section*{Acknowledgments}

The Authors would like to thank all physicians and scientists that provide data to the Polish HLH in Adults Database affiliated by PALG (Polish Adult Leukemia Group).


\ifCLASSOPTIONcaptionsoff
  \newpage
\fi



\bibliographystyle{IEEEtranN}
\bibliography{ieee/bib/IEEEabrv, ieee/tran/bibliografia}
\end{document}